\documentclass{article}

\usepackage[preprint,nonatbib]{neurips_2022}

\usepackage[utf8]{inputenc} % allow utf-8 input
\usepackage[T1]{fontenc}    % use 8-bit T1 fonts
\usepackage{url}            % simple URL typesetting
\usepackage{booktabs}       % professional-quality tables
\usepackage{amsfonts}       % blackboard math symbols
\usepackage{nicefrac}       % compact symbols for 1/2, etc.
\usepackage{microtype}      % microtypography
\usepackage[dvipsnames]{xcolor}
\usepackage{multirow}
\usepackage{graphicx}
\usepackage{floatrow}
\usepackage{amsmath}
\usepackage{amssymb}
\usepackage{arydshln}
\usepackage{adjustbox}
\usepackage{subcaption}
\usepackage{bbding}
\usepackage{xspace}

\usepackage[pagebackref=true,breaklinks=true,colorlinks,bookmarks=false,citecolor=ForestGreen]{hyperref}

\makeatletter
\DeclareRobustCommand\onedot{\futurelet\@let@token\@onedot}
\def\@onedot{\ifx\@let@token.\else.\null\fi\xspace}

\def\eg{\emph{e.g}\onedot}

\makeatother

\makeatletter\renewcommand\paragraph{\@startsection{paragraph}{4}{\z@}
  {.2em \@plus.0ex \@minus.2ex}{-.5em}{\normalfont\normalsize\bfseries}}\makeatother

\title{Multimodal Open-Vocabulary Video Classification via Pre-Trained Vision and Language Models}

\author{
Rui Qian\thanks{Work done during an internship at Google Research.}~~$^{1,2}$\quad 
Yeqing Li$^{1}$\quad
Zheng Xu$^{1}$\quad 
Ming-Hsuan Yang$^{1}$\quad 
Serge Belongie$^{3}$\quad 
Yin Cui$^{1}$\\
\\
$^{1}$Google Research \qquad $^{2}$Cornell University \qquad $^{3}$University of Copenhagen
}

\begin{document}

\maketitle

%%%%%%%%%%%%%%%%%%%%%%%%%%%%%%%%%%%%%%%%%%%%%%%%%%%%%%%%%%%%
\begin{abstract}
Utilizing vision and language models (VLMs) pre-trained on large-scale image-text pairs is becoming a promising paradigm for open-vocabulary visual recognition.
In this work, we extend this paradigm by leveraging motion and audio that naturally exist in video.
We present \textbf{MOV}, a simple yet effective method for \textbf{M}ultimodal \textbf{O}pen-\textbf{V}ocabulary video classification. 
In MOV, we directly use the vision encoder from pre-trained VLMs with minimal modifications to encode video, optical flow and audio spectrogram.
We design a cross-modal fusion mechanism to aggregate complimentary multimodal information.
Experiments on Kinetics-700 and VGGSound show that introducing flow or audio modality brings large performance gains over the pre-trained VLM and existing methods.
Specifically, MOV greatly improves the accuracy on base classes, while generalizes better on novel classes.
MOV achieves state-of-the-art results on UCF and HMDB zero-shot video classification benchmarks, significantly outperforming both traditional zero-shot methods and recent methods based on VLMs.
Code and models will be released.
\end{abstract}

%%%%%%%%%%%%%%%%%%%%%%%%%%%%%%%%%%%%%%%%%%%%%%%%%%%%%%%%%%%%
\section{Introduction}
\label{sec:intro}

Building open-vocabulary models capable of predicting novel visual concepts beyond a fixed set of training classes is of crucial importance in computer vision.
Recently, vision and language models (VLMs) that are jointly trained on large-scale image-text pairs, \eg, CLIP~\cite{radford2021learning} and ALIGN~\cite{jia2021scaling}, demonstrate impressive transferability on a wide range of visual recognition tasks.
Utilizing such strong pre-trained VLMs is becoming a promising paradigm for building open-vocabulary models.
Examples include open-vocabulary object detection~\cite{gu2021open} and image segmentation~\cite{ghiasi2021open, li2022language}.

In this work, we focus on the challenging task of open-vocabulary video classification via pre-trained vision and language models. We set up open-vocabulary video benchmarks by utilizing two existing large-scale video classification datasets: Kinetics-700~\cite{carreira2019short} and VGGSound~\cite{chen2020vggsound}. 
Concretely, we construct two sets of classes: base and novel. For base classes, we have access to both training and testing videos, which aims at helping the pre-trained VLMs adapt to the video domain. While for novel classes, we only have testing videos, mimicking the real-world challenge of open-vocabulary video classification. 

\begin{figure}[t]
 \centering
 \begin{subfigure}[b]{0.23\linewidth}
     \centering
     \includegraphics[width=\textwidth]{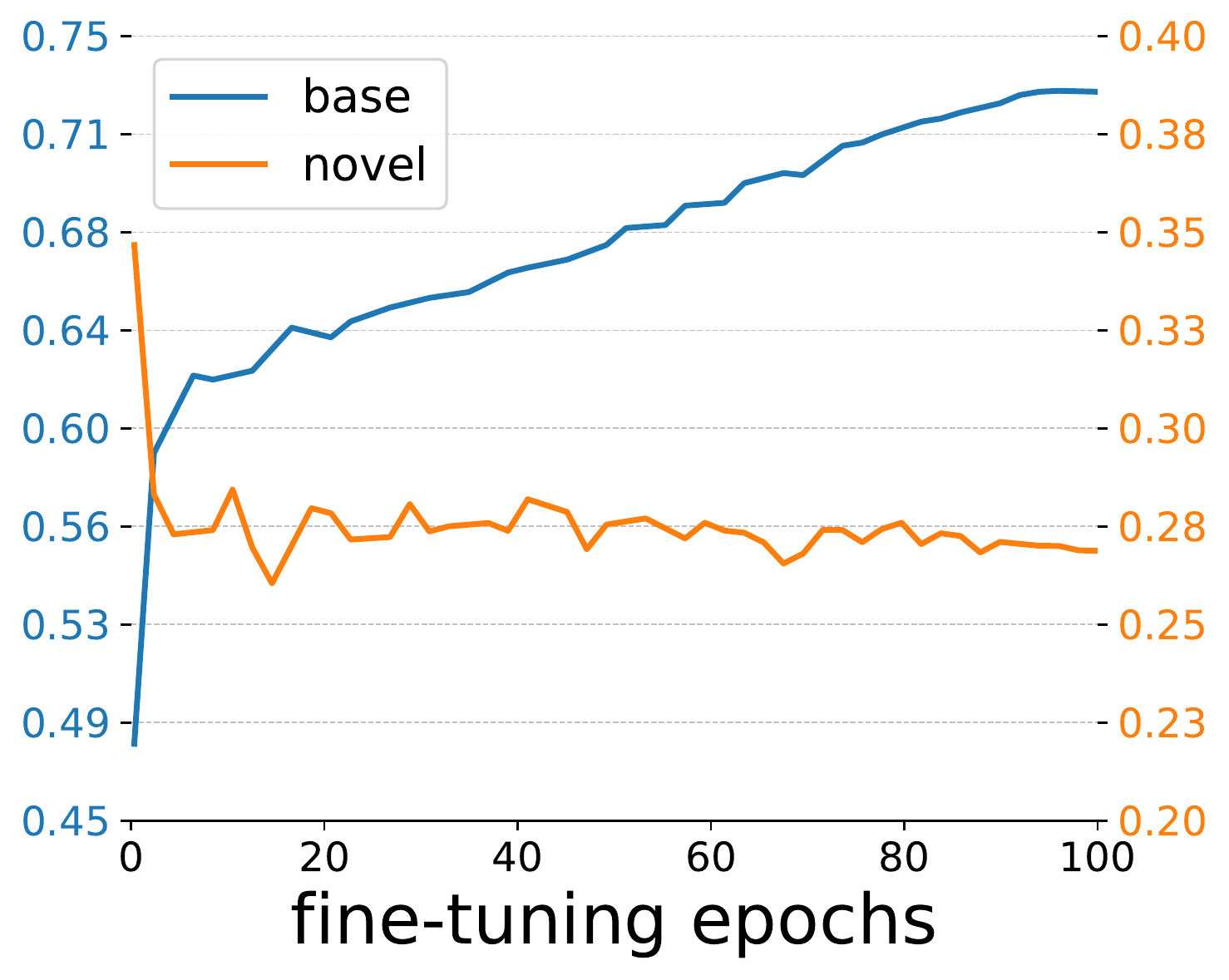}
     \caption{Video in Kinetics}
     \label{fig:modality_generalization_k700_video}
 \end{subfigure}
 \hfill
 \begin{subfigure}[b]{0.23\linewidth}
     \centering
     \includegraphics[width=\textwidth]{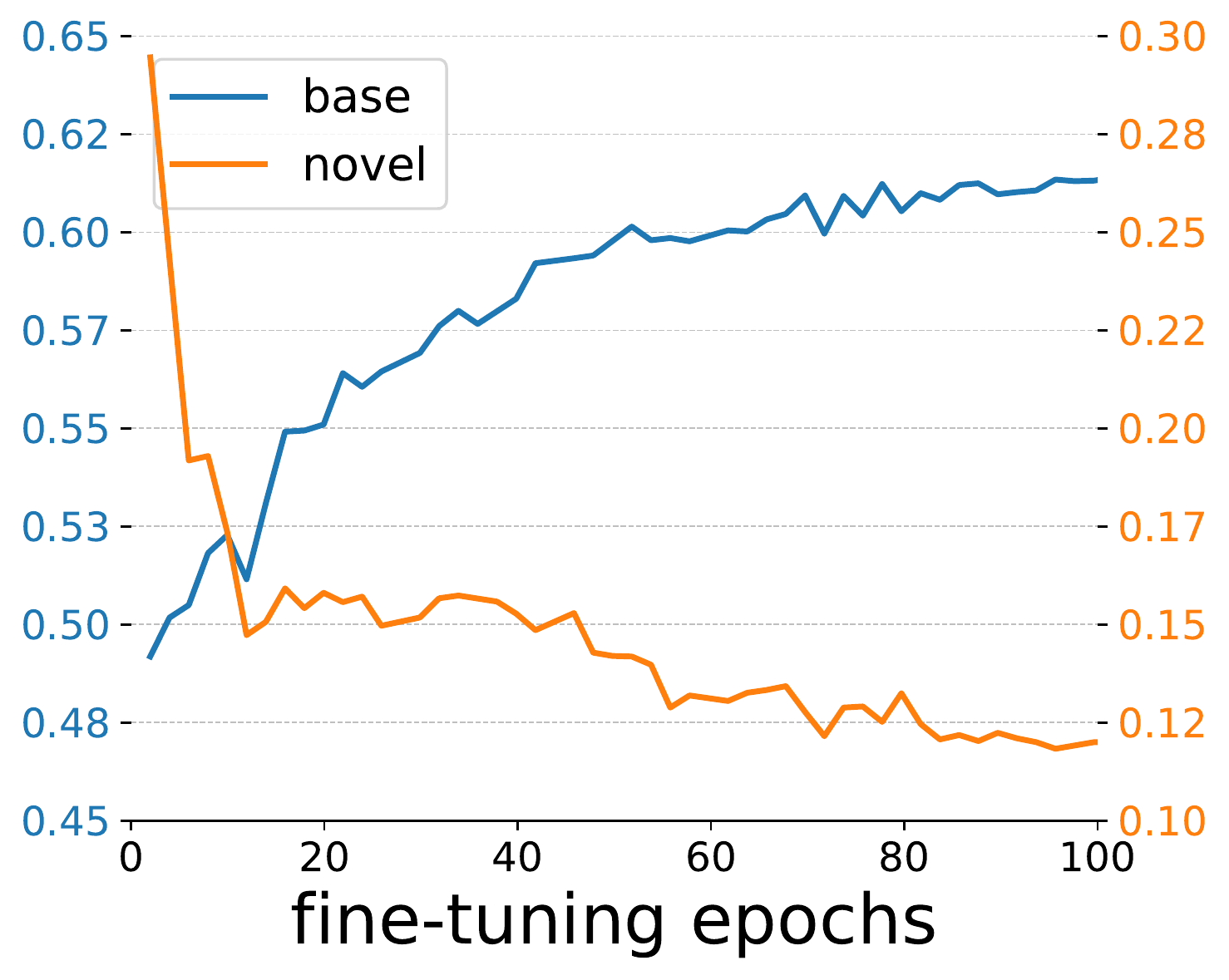}
     \caption{Video in VGGSound}
     \label{fig:modality_generalization_vgg_video}
 \end{subfigure}
 \hfill
 \begin{subfigure}[b]{0.23\linewidth}
     \centering
     \includegraphics[width=\textwidth]{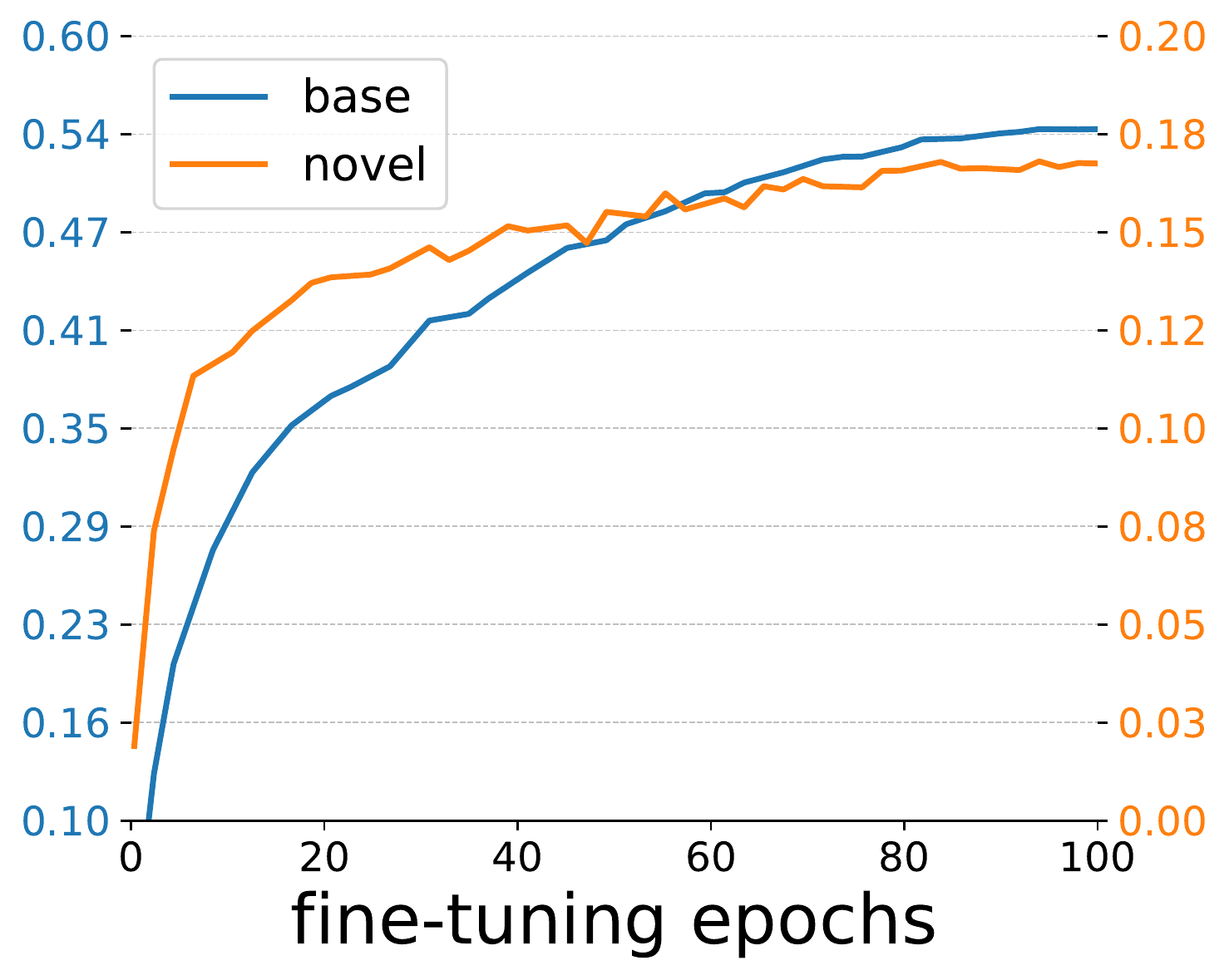}
     \caption{Flow in Kinetics}
     \label{fig:modality_generalization_k700_flow}
 \end{subfigure}
 \hfill
 \begin{subfigure}[b]{0.23\linewidth}
     \centering
     \includegraphics[width=\textwidth]{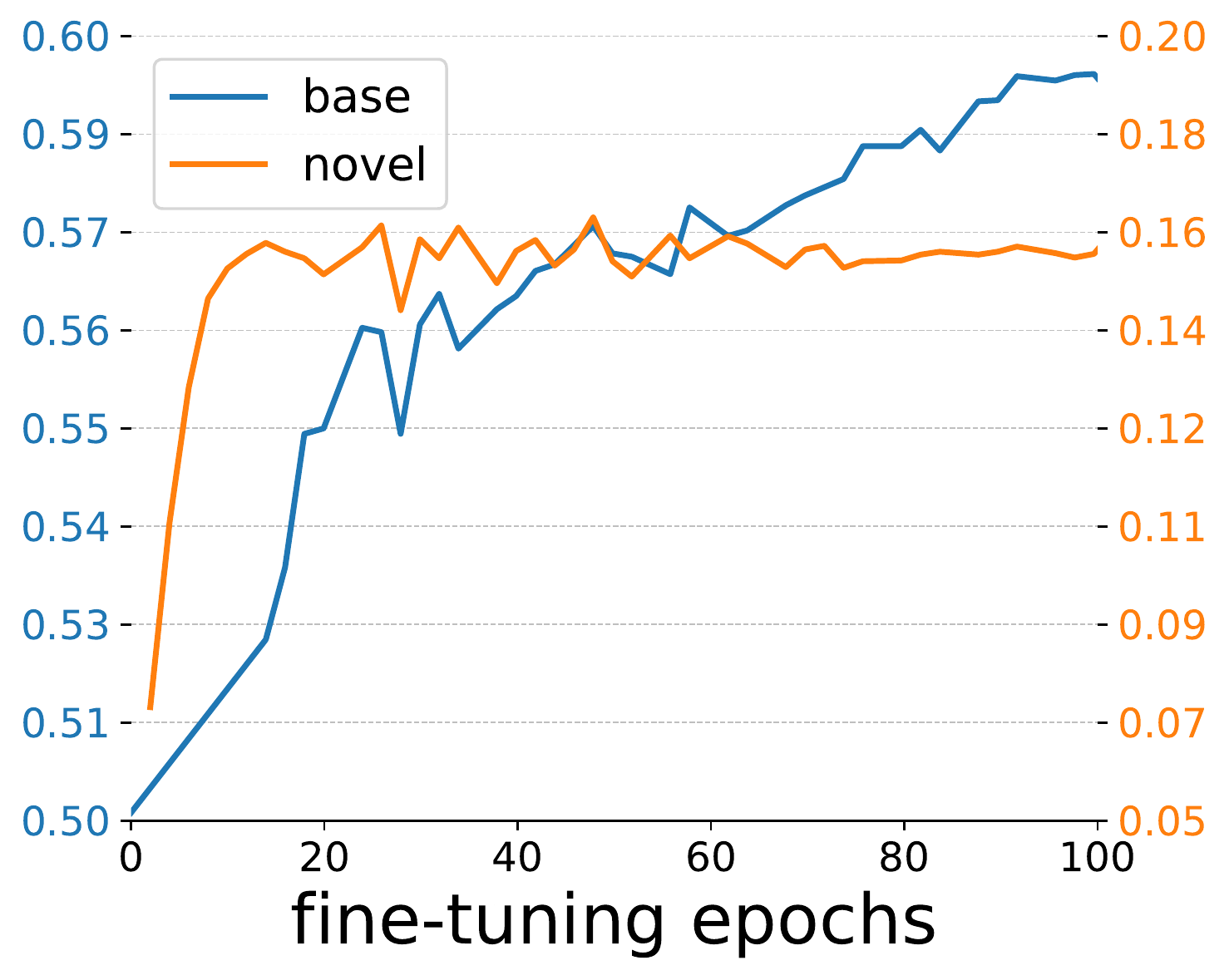}
     \caption{Audio in VGGSound}
     \label{fig:modality_generalization_vgg_audio}
 \end{subfigure}
    \caption{\textbf{Fine-tuning pre-trained CLIP with video, flow and audio modalities.} For all three modalities, fine-tuning on labeled base classes leads to significant accuracy improvement. However, when evaluating the same model on novel classes, the video modality shows decreasing performance, while the performance for both flow and audio modality is improving.}
\label{fig:modality_generalization}
\end{figure}

We start with directly fine-tuning the pre-trained CLIP~\cite{radford2021learning}, a representative vision and language model, using the training videos from base classes. 
As shown in Fig.~\ref{fig:modality_generalization_k700_video} and~\ref{fig:modality_generalization_vgg_video}, we observe that although there is a decent performance improvement on base classes, the accuracy on novel classes decreases significantly.
This finding aligns with some recent work studying the generalization of adapting pre-trained VLMs~\cite{zhou2022cocoop}.

On the other hand, despite the rich multimodal contents in internet videos, signals such as audio and motion are less explored in recent open-vocabulary models.
This is in stark contrast with human perception system that heavily relies on multimodal signals~\cite{smith2005development}.
Can we leverage multimodal information to improve open-vocabulary models?

Instead of using modality-specfic pre-trained encoder networks or methods~\cite{wang2016temporal, hershey2017cnn}, we choose a more straightforward path by directly utilizing the pre-trained vision encoder from VLMs with minimal modifications to deal with optical flow and audio spectrogram. 
Apart from being easy to implement, there are two additional advantages: 1) the vision encoder is pre-trained on large-scale data thus is probably stronger than some commonly used in-domain datasets for pre-training (\eg, ImageNet~\cite{deng2009imagenet}, AudioSet~\cite{gemmeke2017audio}); 2) the vision encoder is trained to align with the language encoder, potentially helping the generalization from base to novel classes.

We fine-tune on the same benchmarks but using flow and audio instead as input modalities.
As shown in Fig.~\ref{fig:modality_generalization_k700_flow} and~\ref{fig:modality_generalization_vgg_audio}, surprisingly, we find that fine-tuning on base classes is able to also improve the performance on novel classes.
This suggests that we may use flow and audio modality to improve the generalization of video modality from base to novel classes.

\begin{figure}[t]
\centering
\includegraphics[width=\textwidth]{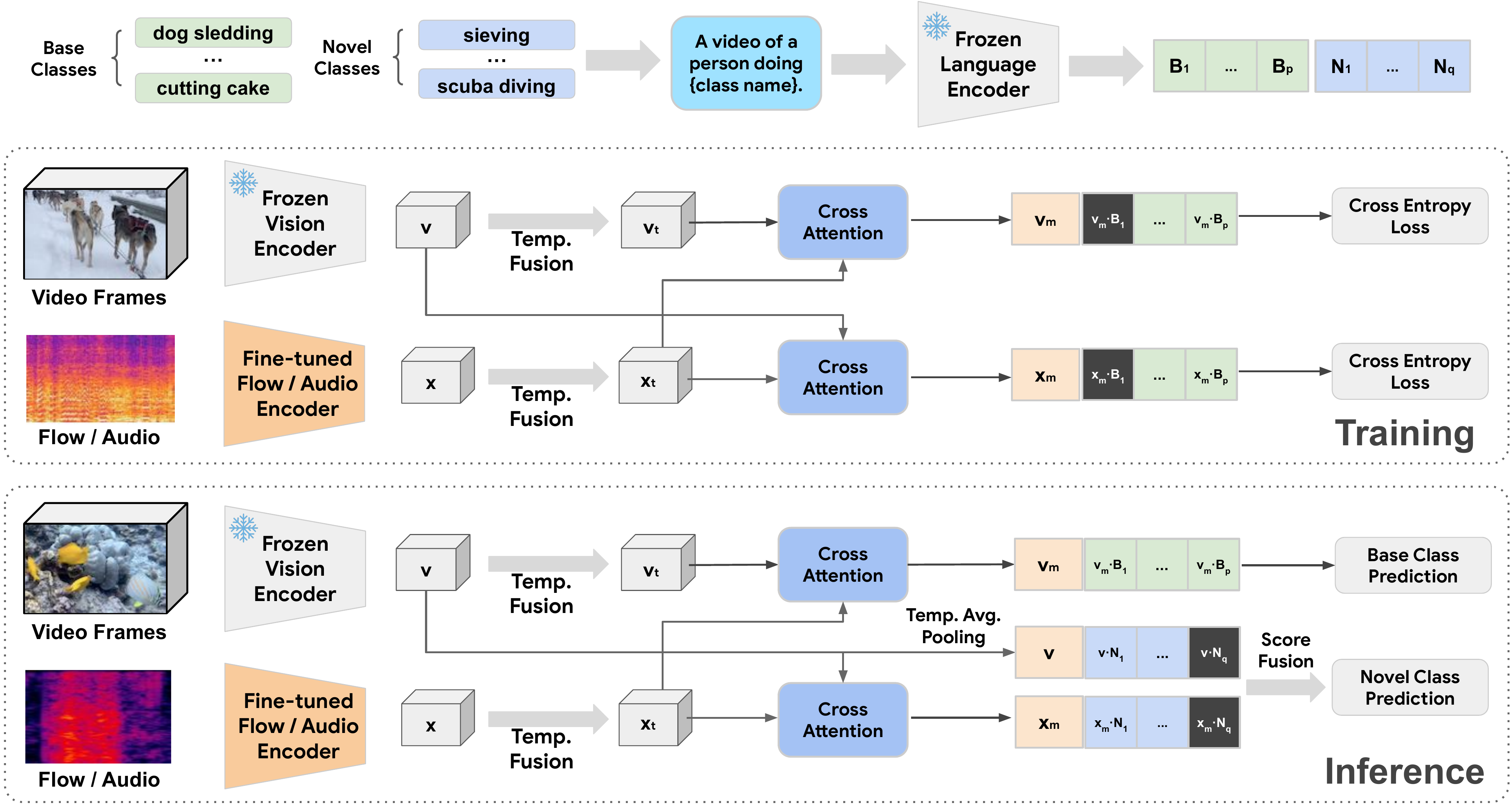}
\caption{\textbf{An overview of the proposed multimodal open-vocabulary (MOV) method}. We use the same encoder architecture from the pre-trained vision and language model to encode the video frames, optical flow and audio spectrogram. We then apply a transformer head for temporal fusion. We design a cross-attention mechanism for fusion across modalities. During training, we optimize different modalities simultaneously via calculating their similarity with the text embeddings. During inference, we use separate paths for base and novel class prediction.}
\label{fig:network}
\end{figure}

In light of our observations, we propose \textbf{MOV}, a simple yet effective method for \textbf{M}ultimodal \textbf{O}pen-\textbf{V}ocabulary video classification.
Fig.~\ref{fig:network} shows an overview of our method.
In MOV, we design a multimodal fusion mechanism using cross-attention to leverage complimentary multimodal information. 
The core idea is to exploit the strong transferability in the pre-trained vision encoder, while allowing greater flexibility in fine-tuning flow and audio encoders.
MOV is trained on multimodal inputs from base classes and is able to predict both base and novel classes during inference.
Sec.~\ref{sec:methods} provides a detailed description of our proposed method.

We conduct extensive experiments and ablation studies on two representative multimodal video datasets: Kinetics-700~\cite{carreira2019short} and VGGSound~\cite{chen2020vggsound}. MOV shows clear improvements over CLIP as well as recent VLM adaptation methods~\cite{zhou2021coop, gao2021clip} on both base and novel classes. 
MOV also achieves state-of-the-art results on UCF and HMDB zero-shot video classification benchmarks, significantly outperforming both traditional zero-shot methods and recent methods based on VLMs. 
Furthermore, MOV is scalable with much stronger backbones, indicating its potentials to be incorporated with giant vision and language models.

%%%%%%%%%%%%%%%%%%%%%%%%%%%%%%%%%%%%%%%%%%%%%%%%%%%%%%%%%%%%
\section{Related work}
\label{sec:related_work}
\paragraph{Vision and language models.} 
Learning a joint embedding space from vision and language modalities has been extensively investigated during the past decade. 
Early works usually first encode two modalities separately, using hand-crafted descriptors~\cite{elhoseiny2013write} or deep networks~\cite{lei2015predicting} for image, and skip-gram text models for language~\cite{frome2013devise}. 
The cross-modality alignment is then achieved by metric learning~\cite{frome2013devise} or language concepts~\cite{li2017learning}. 
Recently, learning vision and language modalities jointly through contrastive learning~\cite{hadsell2006dimensionality, oord2018representation} becomes a promising direction. 
Impressive performance has been achieved by utilizing stronger encoders for vision~\cite{dosovitskiy2020image}, language~\cite{vaswani2017attention} and web-scale pre-training data~\cite{hinton2015distilling, radford2021learning}. 
CLIP~\cite{jia2021scaling} and ALIGN~\cite{radford2021learning} are two representative approaches which shows strong zero-shot~\footnote{We use the term ``zero-shot'' when we need to align with settings described in some existing works. Otherwise, we would use ``open-vocabulary'' which we believe is a more precise term.} performance on various downstream tasks. 
Despite this strong baseline, adapting pre-trained VLMs to specific vision domains in a more effective way remains critical and is being actively studied. 
Examples include image classification~\cite{zhou2021coop, zhou2022cocoop, gao2021clip}, object detection~\cite{gu2021open}, image segmentation~\cite{ghiasi2021open, li2022language} and video action recognition~\cite{wang2021actionclip, ju2021prompting}.
Our method extends the existing research by adapting pre-trained VLMs to multimodal video and investigating the impact of additional input modalities like flow and audio.

\paragraph{Open-vocabulary video classification.} Zero-shot or open-vocabulary video action recognition is a representative task in this domain. 
Similar to early works of vision and language learning, the video input and labeled texts are encoded with modality-specific pre-trained models such as S3D~\cite{xie2018rethinking}, R(2+1)D~\cite{tran2018closer} for video, Word2Vec~\cite{mikolov2013efficient} for text. 
Since the generated video and text embeddings are not aligned, various methods have been proposed to bridge the gap by mapping two modalities into a joint embedding space~\cite{wang2017zero, chen2021elaborative, gao2019know, wu2016harnessing, xu2016multi, zhu2018towards}, mapping vision modality to language space~\cite{bishay2019tarn, brattoli2020rethinking, hahn2019action2vec, xu2017transductive} or mapping language modality to vision space~\cite{mandal2019out, zhang2018visual}.
These joint embedding mapping methods are further extended to audiovisual classification~\cite{mercea2022audio, mazumder2021avgzslnet, parida2020coordinated}.
Our approach shows that we can improve the performance of open-vocabulary video classification by leveraging strong pre-trained VLMs and other modalities like flow and audio.

\paragraph{Mutlimodal fusion for video.} Video is a natural source of multimodal data including motion and audio. Two-stream networks is used to model video and optical flow simultaneously for action classification~\cite{simonyan2014two, wang2016temporal, feichtenhofer2016convolutional, feichtenhofer2017spatiotemporal}. Late fusion is adopted~\cite{simonyan2014two, wang2016temporal} and then thoroughly studied~\cite{feichtenhofer2016convolutional, feichtenhofer2017spatiotemporal} on how to better perform spatio-temporal fusion from two modalities. As in the domain of audiovisual fusion, early methods~\cite{chen1998audio} usually adopt straightforward score fusion or stacking input data for early fusion. Later research~\cite{kazakos2019epic, xiao2020audiovisual, fayek2020large, nagrani2021attention} focus on developing better mid or late fusion strategies to improve the final performance.
Different from existing works focusing on a fixed set of classes, we use multimodal fusion to help open-vocabulary models generalize better to novel classes.

%%%%%%%%%%%%%%%%%%%%%%%%%%%%%%%%%%%%%%%%%%%%%%%%%%%%%%%%%%%%
\section{Methods}
\label{sec:methods}
An overview of our proposed method is shown in Fig.~\ref{fig:network}. We next describe each component.

\subsection{Modality-Specific Encoding}
\label{subsec:mod_encoding}
Given a pre-trained vision and language model, \eg, CLIP \cite{radford2021learning}, we denote its vision encoder as $h(\cdot | \theta_h)$ and its language encoder as $g(\cdot | \theta_g)$. 
For a multimodal video input, we sample $N$ RGB frames $V$ and calculate the corresponding optical flow images $F$, resulting in $V = \{v_1, v_2, \ldots, v_N\}$ and $F = \{f_1, f_2, \ldots, f_N\}$. 
We also generate the spectrogram image $A$ from the raw audio waveform. More implementation details can be found in Sec.~\ref{sec:experiments}. 
We use the same encoder architecture $h(\cdot|\cdot)$ to extract feature representations for video, flow and audio modalities, denoted as $h_v (\cdot | \theta_v)$, $h_f (\cdot | \theta_f)$, and $h_a(\cdot | \theta_a)$ respectively. 
Model parameters $\theta_v$, $\theta_f$ and $\theta_a$ are all initialized with the weight $\theta_h$ from the pre-trained VLM. We encode each modality separately as:
\begin{equation}
    \mathbf{v} = h_v (V | \theta_v),~
    \mathbf{f} = h_f (F | \theta_f),~
    \mathbf{a} = h_a (A | \theta_a),
\end{equation}
where $\mathbf{v}$ and $\mathbf{f}$ are features from $N$ frames, and $\mathbf{a}$ is the representation of a single spectrogram image. 

To better aggregate the temporal features of video and flow modalities, we attach temporal fusion networks $\phi_v(\cdot)$ and $\phi_f(\cdot)$, consisting of $L$ transformer layers each, on top of $h_v (\cdot | \theta_v)$ and $h_f (\cdot | \theta_f)$. We denote the input of the $l$-th transformer layer as $\mathbf{z}^l$ and the input $\mathbf{z}^0$ can be either $\mathbf{v}$ or $\mathbf{f}$. 
Then the forward pass of the $l$-th layer in $\phi_v(\cdot)$ and $\phi_f(\cdot)$ can be formulated as:
\begin{align}
    \mathbf{y}^{l} &= \mathrm{MSA}(\mathrm{LN}(\mathbf{z}^l)) + \mathbf{z}^{l}, \label{eq:encoder_self_attention} \\
    \mathbf{z}^{l + 1} &= \mathrm{MLP}(\mathrm{LN}(\mathbf{y}^l)) + \mathbf{y}^l,
\end{align}
where LN stands for layer normalization, MSA represents multi-head self-attention, and MLP means multi-layer perceptron. 
For audio feature $\mathbf{a}$, we simply attach an MLP module upon the backbone. We obtain the temporally fused features as:
\begin{equation}
    \mathbf{v_t} = \phi_v(\mathbf{v}),~
    \mathbf{f_t} = \phi_f(\mathbf{f}),~
    \mathbf{a_t} = \mathrm{MLP}(\mathbf{a}).
\end{equation}

Finally, for the text modality, suppose we have $p$ base classes with labels. We fill each of the class names into 28 video classification prompts provided by CLIP~\cite{radford2021learning} like ``a video of a person doing $\{$class name$\}$'' and then encode the sentence using the pre-trained language encoder $g(\cdot | \theta_g)$ from VLM. The embedding of each class is averaged over all templates and we denote as $\{\mathbf{B}_i\}_{i=1}^{p}$.

\subsection{Multimodal Fusion}
\label{subsec:mod_fusion}
We adopt a cross-attention mechanism to leverage mutlimodal features. Note that despite we can encode all modalities simultaneously, existing video benchmarks usually only contain two most informative modalities, \eg video and flow for action classification, video and audio for audiovisual classification. Thus our algorithm described here is for fusing one of $\{$flow, audio$\}$ modality with video modality, as shown in Fig.~\ref{fig:network}. 

For the video modality, we extract the information from other modalities to enhance the performance of video feature. Thus we use $\mathbf{v_t}$ as the input for attention query, and $\mathbf{f_t}$ or $\mathbf{a_t}$ from the other modality as the input for attention key and value. The fused multimodal video feature $\mathbf{v_m}$ can be written as:

\begin{align}
    \mathbf{v_t} &= \mathrm{MCA}(\mathrm{LN}(\mathbf{v_t}), \mathrm{LN}(\mathbf{x_t})) + \mathbf{v_t}, ~~\mathbf{x_t} \in \{\mathbf{f_t}, \mathbf{a_t}\},\\
    \mathbf{v_m} &= \mathrm{AvgPool}\big( \mathrm{MLP}(\mathrm{LN}(\mathbf{v_t})) + \mathbf{v_t} \big),
\end{align}
where MCA denotes multi-head cross-attention, AvgPool denotes temporal average pooling.

For the audio and flow modalities, we aim at incorporating the information from video modality to enhance the generalization ability of the feature. Since the parameters of the temporal fusion network $\phi_v(\cdot)$ for generating the video feature $\mathbf{v_t}$ are still trained from scratch on base classes, we choose to directly use the backbone's output $\mathbf{v}$ instead of $\mathbf{v_t}$ for better generalization on novel classes. We obtain the fused multimodal flow and audio feature $\mathbf{f_m}$ and $\mathbf{a_m}$ as:
\begin{align}
    \mathbf{f_t} &= \mathrm{MCA}(\mathrm{LN}(\mathbf{f_t}), \mathrm{LN}(\mathbf{v})) + \mathbf{f_t},~~~~~~~~\mathbf{a_t} = \mathrm{MCA}(\mathrm{LN}(\mathbf{a_t}), \mathrm{LN}(\mathbf{v})) + \mathbf{a_t}, \\
    \mathbf{f_m} &= \mathrm{AvgPool}\big( \mathrm{MLP}(\mathrm{LN}(\mathbf{f_t})) + \mathbf{f_t} \big),~~\mathbf{a_m} = \mathrm{AvgPool}\big( \mathrm{MLP}(\mathrm{LN}(\mathbf{a_t})) + \mathbf{a_t} \big).
\end{align}

\subsection{Training and Inference on Base Classes}
\label{subsec:closeset}
During training, each input multimodal video has a corresponding label $y$ belonging to the base classes. We would optimize different modalities simultaneously via calculating the video-to-text, flow-to-text and audio-to-text similarity. The training loss function can be formulated as:
\begin{equation}
\label{eq:cset_train}
L = \alpha (- \log \frac{\exp (\operatorname{sim} (\mathbf{v_m}, \mathbf{B}_y) / \tau)}{\sum_{i=1}^{p} \exp (\operatorname{sim} (\mathbf{v_m}, \mathbf{B}_i) / \tau)}) + (1 - \alpha) ( - \log \frac{\exp (\operatorname{sim} (\mathbf{x_m}, \mathbf{B}_y) / \tau)}{\sum_{i=1}^{p} \exp (\operatorname{sim} (\mathbf{x_m}, \mathbf{B}_i) / \tau)}),
\end{equation}
where $\mathbf{x_m} \in \{\mathbf{f_m}, \mathbf{a_m}\}$, $\alpha$ is the weight for balancing two loss terms, $\operatorname{sim}(\cdot, \cdot)$ is the cosine similarity, $\tau$ is a pre-defined temperature parameter.  
During training, we freeze the video encoder and the text encoder to save computation and speed up the training, while for the other two modalities flow and audio, we fine-tune the whole encoder end-to-end. 
An ablation study on fine-tuning different number of layers can be found in Tab.~\ref{tab:ablation_ft_layer}.

For inference on base classes, we compute the probability belonging to the $j$-th class by:
\begin{equation}
\label{eq:cset_pred}
  P(j) = \frac{\exp (\operatorname{sim} (\mathbf{v_m}, \mathbf{B}_j) / \tau)}{\sum_{i=1}^{p} \exp (\operatorname{sim} (\mathbf{v_m}, \mathbf{B}_i) / \tau)},~j \in \{1, 2, \ldots, p\}.
\end{equation}

\subsection{Generalization to Novel Classes}
\label{subsec:openset}
Similar to base classes, we obtain the text embeddings for novel classes as $\{\mathbf{N}_i\}_{i=1}^{q}$, where $q$ is the number of novel classes. 
In addition to fused features $\mathbf{f_m}$ or $\mathbf{a_m}$, we also incorporate the video feature $\mathbf{v}$ extracted from the frozen video backbone, followed by a temporal average pooling. Similar to Eq.~\ref{eq:cset_pred}, we compute the probability predictions as (here we only show flow modality for simplicity): 
\begin{equation}
\label{eq:oset_pred}
P_{f}(j) = \frac{\exp (\operatorname{sim} (\mathbf{f_m}, \mathbf{N}_j) / \tau_f)}{\sum_{i=1}^{q} \exp (\operatorname{sim} (\mathbf{f_m}, \mathbf{N}_i) / \tau_f)},~P_{v}(j) = \frac{\exp (\operatorname{sim} (\mathbf{v}, \mathbf{N}_j) / (\tau_v)}{\sum_{i=1}^{q} \exp (\operatorname{sim} (\mathbf{v}, \mathbf{N}_i) / (\tau_v))},~j \in \{1, 2, \ldots, q\}.
\end{equation}
We denote the probability distribution followed by $\{p_{f}(j) |_{j=1}^q\}$ and $\{p_{v}(j) |_{j=1}^q\}$ as $D_f$ and $D_v$. 
In our experiments we find the curve of $D_v$ tends to be much flatter (or have higher information entropy) than $D_f$ when the temperatures $\tau_v$ and $\tau_f$ are both set to the CLIP's default value of 0.01, resulting in poor performance.
We find simply setting $\tau_v$ to 0.003 while keeping $\tau_f$ and $\tau_a$ as 0.01 solves this issue.
A detailed ablation study about the temperature can be found in Appendix~\ref{appx:temperature_tuning}.

The final probability predictions for novel classes are calculated by a weighted sum:
\begin{equation}
\label{eq:oset_fusion}
P(j) = \beta P_{f}(j) + (1 - \beta )P_{v}(j).
\end{equation}

%%%%%%%%%%%%%%%%%%%%%%%%%%%%%%%%%%%%%%%%%%%%%%%%%%%%%%%%%%%%
\section{Experiments}
\label{sec:experiments}

\subsection{Data}
\label{subsec:data_prep}

We describe the details of dataset splits for benchmarking multimodal open-vocabulary video classification and preparing flow and audio modalities. 

\paragraph{Kinetics-700~\cite{carreira2019short} splits.} 
Kinetics-700 contains around 650k video clips annotated with 700 human action classes. Apart from the visual modality, the optical flow modality plays an important role for distinguishing different action classes. 
For dataset split, we randomly select 400 classes as base classes and the testing videos of the rest 300 classes are used for novel classes evaluation. 

\paragraph{Kinetics-700 optical flow.} 
We follow a standard procedure~\cite{xie2018rethinking, han2020memory, han2020self} to use the TV-L1 algorithm~\cite{zach2007duality} to extract optical flow in an unsupervised manner. To accommodate for pre-trained vision encoders, we first truncate the vertical and horizontal motion values to $[-20, 20]$, then append a third all-zero channel. Finally we do a shift and scale transformation to map $[-20, 20]$ to $[0, 255]$.  

\paragraph{VGGSound~\cite{chen2020vggsound} splits.} VGGSound contains around 200k video clips belonging to a total number of 309 classes. Different from other audiovisual datasets like AudioSet~\cite{gemmeke2017audio}, VGGSound ensures the source of the sound is visually present inside the same video. Thus we consider this dataset as an excellent test bed for our proposed method. 
We randomly select 154 base classes for training and leave the rest 155 classes for novel classes evaluation.

\paragraph{VGGSound audio spectrogram.} 
We follow the pre-processing practice of audio spectrogram transformer (AST)~\cite{gong21ast} to convert wavforms to spectrogram images. First, the raw audio signal is re-sampled to 16kHz and converted to mono channel. We then calculate the log mel spectrogram with 128 frequency bins. The processing Hamming window is 25ms with a hop length set to 10ms. For \emph{t} second audio input, the generated 2D spectrogram would have the shape of $128 \times 100\emph{t}$. We normalize the spectrogram by subtracting the mean pixel value and dividing the standard deviation. 

\subsection{Implementation}
\label{subsec:detail}
\paragraph{Data augmentation and tokenization.} For video, we first randomly sample 16 frames with a stride of 4 from the whole video sequence. We then apply the standard image augmentation used on ImageNet~\cite{he2016deep, he2019bag} with same augmentation parameters across all frames to keep temporal consistency~\cite{qian2021spatiotemporal}. 
For optical flow, we follow the practice of~\cite{xie2018rethinking, han2020memory, han2020self} by directly treating it as images and apply the same augmentation with the video. The augmentated output tensors have the shape of $(16, 224, 224, 3)$ from both modalities which can be directly fed into CLIP's vision encoder~\cite{radford2021learning}. 
For audio, we apply specialized augmentations designed for spectrogram following~\cite{gong21ast, nagrani2021attention}. As the videos in VGGSound are all 10-second long, the generated spectrogram has a shape of $(128, 100\times10)$. We first conduct a random cropping of $(128, 800)$, sampling all frequency bands with a time duration of 8 seconds. 
SpecAugment~\cite{park2019specaugment} is applied subsequently with a time masking range of 192 frames and frequency masking of 48 bins. 
Finally, to accomodate this single channel output with the pre-trained tokenization layer, we make two necessary changes following~\cite{gong21ast}: 1) expanding the spectrogram to three duplicated channels, 2) bilinearly interpolating the original positional encoding for spectrogram images with a different resolution.  

\paragraph{Network architecture.} We adopt CLIP's ViT-B/16 encoder for video, flow, and audio and the transformer encoder for text. We stack 2 transformer layers for temporal fusion, with an embedding dimension of 512 and 8 attention heads. For the cross-attention head, we use 1 transformer decoder layer with 8 attention heads and an embedding dimension of 512. Query and key-value inputs use separate layer normalization. 

\paragraph{Training hyper-parameters.} We set all hyper-parameters except for train epochs same for experiments on Kinetics-700 and VGGSound. We use a batch size of 1024 on 128 Cloud TPUv3 cores, AdamW~\cite{loshchilov2017decoupled} optimizer with a weight decay of 0.05 and an initial learning rate of 1e-4 followed by half-cosine decay~\cite{he2019bag}. We set the weight $\alpha$ in Eq.~\ref{eq:cset_train} as 0.5. We train 100 epochs on Kinetics-700 and 20 epochs on VGGSound since we observe an overfitting issue with audio modality when trained longer. 

\paragraph{Inference hyper-parameters.}
During inference, for video and flow, we use $4\times3$ views following~\cite{arnab2021vivit, liu2021video} where a video is uniformly sampled into 4 clips temporally, and 3 spatial crops are conducted for each clip. 
For audio, we use 12 temporal views without spatial cropping. The final score is averaged over 12 views. For novel classes, we set the weight $\beta$ in Eq.~\ref{eq:oset_fusion} to 0.25.

\subsection{Multimodal Open-Vocabulary Video Classification} 
We evaluate MOV on Kinetics-700 to utilize modalities of video, optical flow and text, and on VGGSound to explore the combination of video, audio and text. 

\paragraph{Comparison baselines.} We compare with three baselines: \textbf{1)} CLIP~\cite{radford2021learning}, which directly encodes the video and class names into embeddings with pre-trained encoders. The final prediction is given by comparing similarity scores between video and text embeddings; 
\textbf{2)} CoOp~\cite{zhou2021coop}, which learns continuous text prompt embeddings instead of manually selected templates for better adaptation to downstream tasks; 
\textbf{3)} CLIP-Adapter~\cite{gao2021clip}, which attaches adapter heads to both video and text encoder. We use the same data, backbone and hyper-parameters as ours introduced in Sec.~\ref{subsec:detail} to train (CLIP doesn't require training) and evaluate all methods.

\paragraph{Results.} 
Tab.~\ref{tab:exp_kinetics700} shows results on Kinetics-700. 
We can see that both CoOp and CLIP-Adapter achieve better performance than CLIP on base class prediction. While for novel classes, we observe a large accuracy drop compared with CLIP. 
The worse performance in harmonic mean of these two methods indicates the loss of the generalization ability outweigh their improvements on base classes. 
Our proposed MOV shows better performance on base classes, demonstrating the effectiveness of multimodal fusion. On novel classes, we also observe an improvement of 1.4\% over CLIP, indicating that bringing in flow modality improves the generalization of the open-vocabulary model.

\begin{table}[t]
\begin{tabular}{llccc}
    \toprule 
    method & modalities & base acc. & novel acc. & harmonic mean      \\
    \midrule 
    CLIP~\cite{radford2021learning} & V, T & 51.2 & 56.7 & 53.8 \\
    CoOp~\cite{zhou2021coop} & V, T & 58.9 & 45.7 & 51.5 \\
    CLIP-Adapter~\cite{gao2021clip} & V, T & 66.5 & 36.2 & 46.9 \\
    MOV & V, F, T & \textbf{75.3} & \textbf{58.1} & \textbf{65.6}\\
    \bottomrule 
\end{tabular}
\caption{\textbf{Open-vocabulary video classification on Kinetics-700~\cite{carreira2019short}}. Modalities are V: Vision, F: Optical Flow and T: Text. MOV obtains the best performance on both base and novel classes, surpassing CLIP~\cite{radford2021learning} by 24.1 and 1.4, respectively.}
\label{tab:exp_kinetics700}
\end{table}

We observe similar trends in experiments on VGGSound in Tab.~\ref{tab:exp_vggsound}. 
CoOp and CLIP-Adapter gain improvement in base classes but fail to generalize to novel classes, resulting in a lower harmonic mean of accuracy compared to the CLIP baseline.
It is worth noticing that MOV, when fused with the rich audio modality information, shows a 2.7\% improvement on novel classes compared with CLIP. 

\begin{table}[t]
\begin{tabular}{llccc}
    \toprule 
    method & modalities & base acc. & novel acc. & harmonic mean  \\
    \midrule 
    CLIP~\cite{radford2021learning} & V, T & 48.5 & 48.8 & 48.6 \\
    CoOp~\cite{zhou2021coop} & V, T & 56.9 & 42.0 & 48.3 \\
    CLIP-Adapter~\cite{gao2021clip} & V, T & 60.0 & 27.5 & 37.7 \\
    MOV & V, A, T & \textbf{68.4} & \textbf{51.5} & \textbf{58.8} \\
    \bottomrule 
\end{tabular}
\caption{\textbf{Open-vocabulary video classification on VGGSound~\cite{chen2020vggsound}}. Modalities are V: Vision, A: Audio and T: Text. MOV achieves the best performance on both base and novel classes, surpassing CLIP~\cite{radford2021learning} by 19.9 and 2.7, respectively. }
\label{tab:exp_vggsound}
\end{table}

\paragraph{Backbone scaling.} 
It is also important to investigate the scalability of MOV with stronger backbones. We experiment with the largest ViT-L/14 model released by CLIP as the vision encoder and a text encoder with embedding dimension increased to 768 and attention heads increased to 12. 
ViT-L/14 contains 3$\times$ more parameters than ViT-B/16 and we observe around 8\% improvement on direct CLIP zero-shot evaluation on Kinetics-700 and 5\% improvement on VGGSound, as indicated by results in row 1 and 3 in Tab.~\ref{tab:exp_modal_scaling}. 
Despite this much stronger baseline, MOV still improves 20.5\% and 1.6\% on Kinetics-700, 19.3\% and 2.0\% on VGGSound, when compared with row 3 and 4. 
The nice scaling performance shows that MOV has a great potential to be incorporated into recent giant vision and language models~\cite{yuan2021florence, yu2022coca}.  

\begin{table}[hbtp]
\begin{tabular}{lccccc}
\toprule
\multirow{2}{*}{method} & \multirow{2}{*}{backbone} & \multicolumn{2}{c}{Kinetics-700} & \multicolumn{2}{c}{VGGSound} \\
& & base acc. & novel acc. & base acc. & novel acc. \\
\midrule
CLIP~\cite{radford2021learning} & ViT-B/16 & 51.2 & 56.7 & 48.5 & 48.8 \\
MOV & ViT-B/16 & 75.3 & 58.1 & 68.4 & 51.5 \\
\hdashline
CLIP~\cite{radford2021learning} & ViT-L/14 & 59.6 & 65.3 & 52.6 & 54.1 \\
MOV & ViT-L/14 & \textbf{80.1} & \textbf{66.9} & \textbf{71.9} & \textbf{56.1} \\
\bottomrule         
\end{tabular}
\caption{\textbf{Scalability of MOV}. MOV scales well with a stronger ViT-L/14 backbone.}
\label{tab:exp_modal_scaling}
\end{table}

\subsection{Cross-Dataset Transfer}
Pre-training an open-vocabulary or zero-shot video classifcation model on large datasets like Kinetics~\cite{carreira2019short}, ImageNet~\cite{deng2009imagenet} or Sports-1M~\cite{karpathy2014large} and evaluating on UCF101~\cite{soomro2012ucf101} and HMDB51~\cite{kuehne2011hmdb} is the most common paradigm in the literature. 
Following~\cite{brattoli2020rethinking}, there are two major evaluation settings. The first is randomly choosing half of the test dataset’s classes and evaluate on the selected subset. To avoid fluctuations brought by randomness, the evaluation is conducted independently for 10 times and we report the mean accuracy with standard deviation from all trials. 
We donate this setting as UCF$^\dagger$ and HMDB$^\dagger$ in Tab.~\ref{tab:exp_cross_dataset}. 
The second evaluation setting is directly evaluating on the whole dataset, which is suitable for methods pre-trained purely on other datasets~\cite{brattoli2020rethinking, wang2021actionclip, lin2022cross}. We train MOV only using 400 base classes subsampled from Kinectis-700, with video, flow and text. For evaluating on UCF and HMDB, we also use the same three modalities. The flow processing follows the same procedure described in Sec.~\ref{subsec:data_prep}. 

We present detailed and comprehensive comparisons in Tab.~\ref{tab:exp_cross_dataset}, following~\cite{lin2022cross} we list the encoder architecture, pre-train data used and the text encoding method.
Compared with previous state-of-the-art zero-shot video classification approaches (listed above the dashed line), methods adopting pre-trained vision and language model like CLIP obtain much stronger performance. 
Overall, MOV shows clear improvement upon the CLIP baseline with around 3\% on UCF101 for both random sampling and full evaluation, and around 6\% on HMDB51.
Compared with recently proposed ActionCLIP which fine-tunes both vision and text encoder, MOV performs 6.7\% better on UCF101 and 1.6\% better on HMDB51.

\begin{table}[t]
\centering
\footnotesize	
\begin{tabular}{lcccrr}
\toprule 
method & encoder & pre-train data & text & UCF$^\dagger$ / UCF & HMDB$^\dagger$ / HMDB \\
\midrule
GA~\cite{mishra2018generative} & C3D~\cite{tran2015learning} & S1M~\cite{karpathy2014large} & W2V~\cite{mikolov2013efficient} & 17.3$\pm$1.1 / ~~-~~~ &  19.3$\pm$2.1 / ~~-~~~ \\
TARN~\cite{bishay2019tarn} & C3D~\cite{tran2015learning} & S1M~\cite{karpathy2014large} & W2V~\cite{mikolov2013efficient} & 19.0$\pm$2.3 / ~~-~~~ & 19.5$\pm$4.2 / ~~-~~~ \\
CWEGAN~\cite{mandal2019out} & I3D~\cite{carreira2017quo} & IN, K400~\cite{kay2017kinetics} & W2V~\cite{mikolov2013efficient} & 26.9$\pm$2.8 / ~~-~~~ & 30.2$\pm$2.7 / ~~-~~~ \\
TS-GCN~\cite{gao2019know} & GLNet~\cite{szegedy2015going} & IN-shuffle~\cite{mettes2016imagenet} & W2V~\cite{mikolov2013efficient} & 34.2$\pm$3.1 / ~~-~~~ & 23.2$\pm$3.0 / ~~-~~~ \\
PS-GNN~\cite{gao2020learning} & GLNet~\cite{szegedy2015going} & IN-shuffle~\cite{mettes2016imagenet} & W2V~\cite{mikolov2013efficient} & 36.1$\pm$4.8 / ~~-~~~ & 25.9$\pm$4.1 / ~~-~~~ \\
E2E~\cite{brattoli2020rethinking} & R(2+1)D~\cite{tran2018closer} & K700~\cite{carreira2019short} & W2V~\cite{mikolov2013efficient} & ~~48.0 / 35.3 & ~~32.7 / 24.8 \\
DASZL~\cite{kim2021daszl} & TSM~\cite{lin2019temporal} & IN, K400~\cite{kay2017kinetics} & Attributes & 48.9$\pm$5.8 / ~~-~~~ & ~~-~~~ / ~~-~~~ \\
ER~\cite{chen2021elaborative} & TSM~\cite{lin2019temporal} & IN, K400~\cite{kay2017kinetics} & BERT-ED & 51.8$\pm$2.9 / ~~-~~~ & 35.3$\pm$4.6 / ~~-~~~ \\
ResT~\cite{lin2022cross} & RN101~\cite{he2016deep} & K700~\cite{carreira2019short} & W2V~\cite{mikolov2013efficient} & 58.7$\pm$3.3 / 40.6 & 41.1$\pm$3.7 / 34.4 \\
\hdashline
CLIP~\cite{radford2021learning} & ViT-B/16~\cite{dosovitskiy2020image} & Web~\cite{radford2021learning} & TSF~\cite{vaswani2017attention} & 79.9$\pm$3.8 / 73.0 & 54.0$\pm$4.1 / 46.1 \\
ActionCLIP~\cite{wang2021actionclip} & ViT-B/16~\cite{dosovitskiy2020image} & Web~\cite{radford2021learning} & TSF~\cite{vaswani2017attention} & ~~-~~~~~~ / 69.5 & ~~-~~~~~~  / 50.5 \\
MOV & ViT-B/16~\cite{dosovitskiy2020image} & Web~\cite{radford2021learning} & TSF~\cite{vaswani2017attention} & \textbf{82.6$\pm$4.1} / \textbf{76.2} & \textbf{60.8$\pm$2.8} / \textbf{52.1}  \\
MOV & ViT-L/14~\cite{dosovitskiy2020image} & Web~\cite{radford2021learning} & TSF~\cite{vaswani2017attention} & \textbf{87.1$\pm$3.2} / \textbf{80.9} & \textbf{64.7$\pm$3.2} / \textbf{57.8}  \\
\bottomrule 
\end{tabular}
\caption{\textbf{Cross-dataset transfer on UCF and HMDB}. We directly evaluate our proposed MOV without any additional training on two classic video action classification benchmarks. In pre-train data, IN is in short for ImageNet. For the text, BERT-ED means BERT~\cite{devlin2018bert} encoding of elaborated descriptions collected from Wiki/Diction./WordNet. MOV shows the best performance compared with classic zero-shot video classification methods, as well as CLIP and ActionCLIP, demonstrating a strong cross-dataset generalization ability.}
\label{tab:exp_cross_dataset}
\end{table}

\subsection{Ablation Study}
\paragraph{Single modality and fusion.} 
We conduct experiments with single modality to understand the capability of each modality as well as the relative improvement brought by different fusion strategies. 
Results are in Tab.~\ref{tab:exp_single_modality}. 
For Kinetics-700, simply using the optical flow as input obtains 54.2\% on base classes and 16.8\% on novel classes. 
When using score fusion, compared with video modality, we observe an improvement of 1.2\% on novel classes but identical performance on base classes. 
Equipped with the proposed cross-attention fusion mechanism, we obtain 2.6\% improvement on base classes, and 3.5\% on novel classes. 
For VGGSound, the perfromance of audio only is quite close to video only and the score fusion works quite well for base classes with a 6.5\% improvement. Cross-attention further improves the score fusion by 0.7\% in base classes and 2.0\% on novel classes.

\begin{table}[hbtp]
\begin{minipage}[t]{0.45\textwidth}
\centering
\begin{tabular}{lcc}
    \toprule 
    & base acc. & novel acc. \\
    \midrule 
    Flow only & 54.2 & 16.8 \\
    Video only & 72.7 & 26.9 \\
    \hdashline
    Score fusion & 72.7 & 28.1\\
    Cross-attention & \textbf{75.3} & \textbf{30.4}\\
    \bottomrule 
\end{tabular}
\end{minipage}
\hfill
\begin{minipage}[t]{0.45\textwidth}
\centering
\begin{tabular}{lcc}
    \toprule 
    & base acc. & novel acc. \\
    \midrule 
    Audio only & 59.5 & 15.9\\
    Video only & 61.2 & 21.5 \\
    \hdashline
    Score fusion & 67.7 & 22.8\\
    Cross-attention & \textbf{68.4} & \textbf{24.8}\\
    \bottomrule 
\end{tabular}
\end{minipage}

\begin{minipage}[b]{0.45\textwidth}
\vspace{3pt}
\hspace{-40pt}
\centering
(a) \textbf{Kinectics-700.}
\end{minipage}
\begin{minipage}[b]{0.45\textwidth}
\vspace{3pt}
\hspace{40pt}
\centering
(b) \textbf{VGGSound.}
\end{minipage}

\caption{\textbf{Ablation on multimodal fusion}. Mutlimodal fusion improves the perfromance of using single modality, and the proposed cross-attention mechanism works better than score fusion.}
\label{tab:exp_single_modality}
\end{table}

\paragraph{Fine-tuning.} We ablate fine-tuning different layers of the encoder for flow and audio modality and show results in Tab.~\ref{tab:ablation_ft_layer}. As mentioned in Sec.~\ref{sec:methods}, we use the same ViT-B/16 encoder and same initialization weight for video, flow and audio. We iterate choices of fine-tuning the last 1, 3, 6, 9, and all 12 layers and find the performance increases with increasing number of trainable layers on both modalities. Therefore we adopt the setting of fine-tuning all layers for flow and audio modality.

\begin{table}[hbtp]
\begin{minipage}[t]{0.45\textwidth}
\centering
\begin{tabular}{lcc}
    \toprule 
    trainable layers & modality & accuracy \\
    \midrule 
    All 12 layers & Flow & \textbf{54.2} \\
    Last 9 layers & Flow & 51.6 \\
    Last 6 layers & Flow & 46.0 \\
    Last 3 layers & Flow & 38.3 \\
    Last 1 layers & Flow & 30.5 \\
    \bottomrule 
\end{tabular}
\end{minipage}
\hfill
\begin{minipage}[t]{0.45\textwidth}
\centering
\begin{tabular}{lcc}
    \toprule 
    trainable layers & modality & accuracy \\
    \midrule 
    All 12 layers & Audio & \textbf{59.5}\\
    Last 9 layers & Audio & 57.1 \\
    Last 6 layers & Audio & 50.8 \\
    Last 3 layers & Audio & 47.8 \\
    Last 1 layers & Audio & 40.1 \\
    \bottomrule 
\end{tabular}
\end{minipage}

\begin{minipage}[b]{0.45\textwidth}
\vspace{3pt}
\hspace{-40pt}
\centering
(a) \textbf{Fine-tuning on Kinectics-700.}
\end{minipage}
\begin{minipage}[b]{0.45\textwidth}
\vspace{3pt}
\hspace{40pt}
\centering
(b) \textbf{Fine-tuning on VGGSound.}
\end{minipage}

\caption{\textbf{Ablation on fine-tuning different vision encoder layers}. We report the performance on base classes of both datasets, and we find the best setting is fine-tuning all layers of the vision encoder.}
\label{tab:ablation_ft_layer}
\end{table}

\begin{figure}[hbtp]
 \centering
 \begin{subfigure}[b]{0.46\linewidth}
     \centering
     \includegraphics[width=\textwidth]{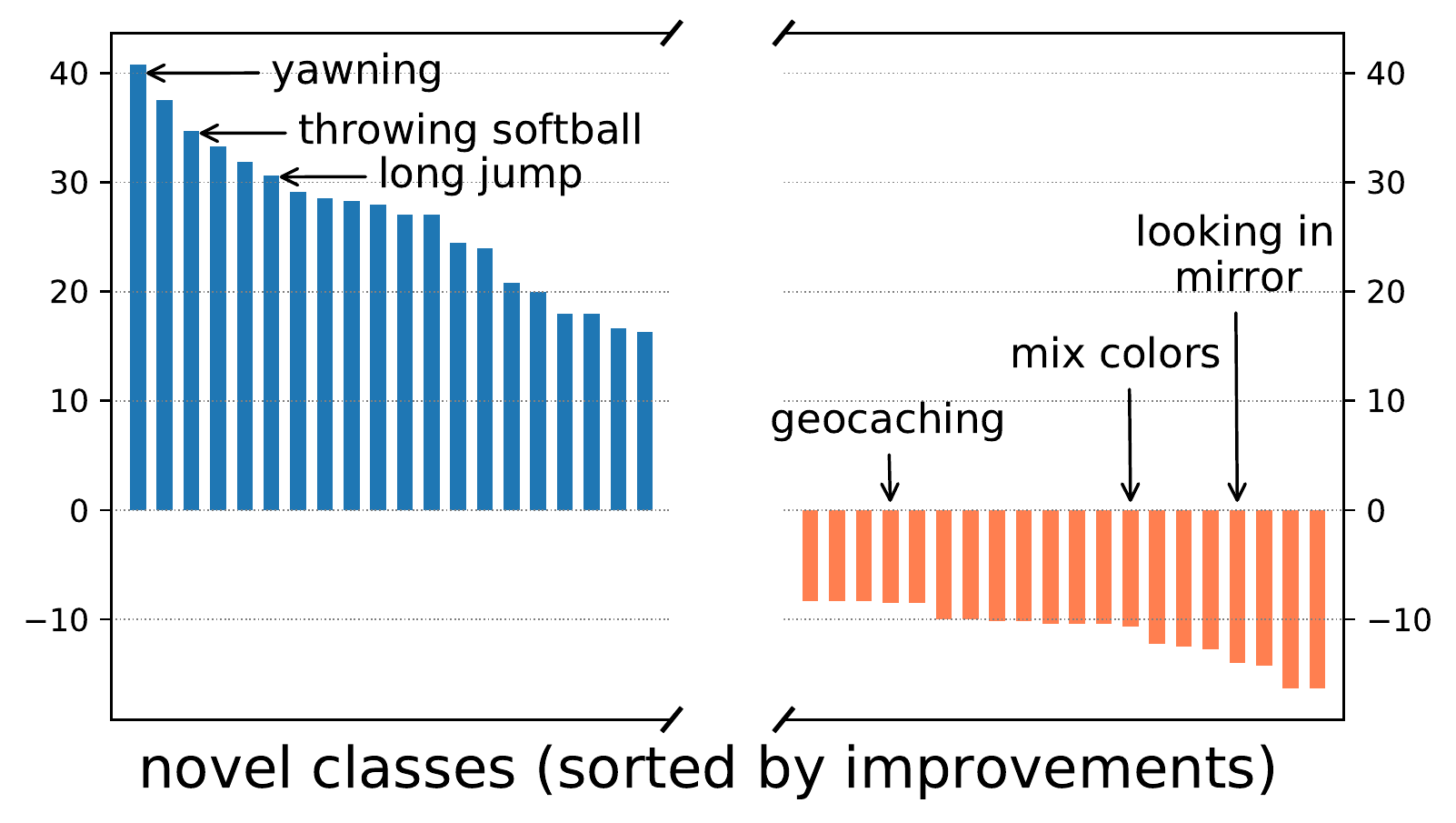}
     \caption{
     \textbf{Kinetics with additional flow modality}}
     \label{fig:k700_per_class}
 \end{subfigure}
 \hfill
 \begin{subfigure}[b]{0.46\linewidth}
     \centering
     \includegraphics[width=\textwidth]{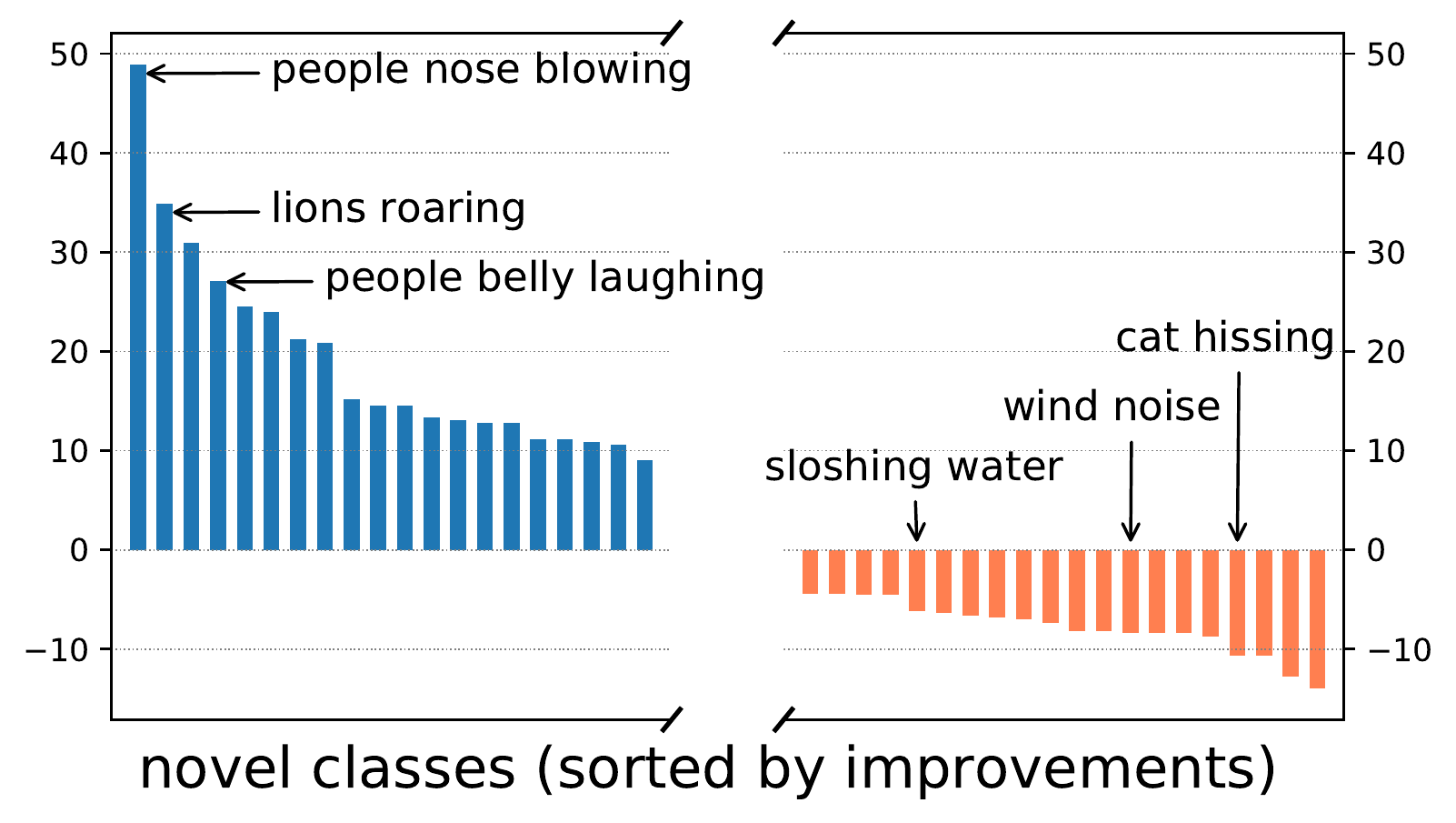}
     \caption{\textbf{VGGSound with additional audio modality}}
     \label{fig:vggsound_per_class}
 \end{subfigure}
    \caption{\textbf{Per-class improvement analysis.} We show top 20 classes with the most improvement (\%) and top 20 classes with the most degradation (\%) when compare the proposed MOV with CLIP.}
\label{fig:per_class_acc}
\end{figure}

\paragraph{Per-class accuracy analysis.} 
We analyze and interpret class-wise performance difference between MOV and CLIP baseline which only uses video and text. 
As illustrated in Fig.~\ref{fig:k700_per_class}, we observe strong gains on classes that require motion understanding, \eg yawning and long jump. While we also find decreased performance on classes with subtle or ambiguous motions, \eg look in mirror and geocaching. 
In Fig.~\ref{fig:vggsound_per_class}, we observe audio modality can significantly help disambiguate classes sharing similar visual contents, \eg people nose blowing and people laughing. While for classes being difficult in the audio domain, \eg sloshing water and wind noise, the performances are degraded.

%%%%%%%%%%%%%%%%%%%%%%%%%%%%%%%%%%%%%%%%%%%%%%%%%%%%%%%%%%%%
\section{Conclusion}
We propose a multimodal open-vocabulary video classification method named MOV via adopting pre-trained vision and language models. Our method is motivated by the observation of drastic performance difference when using video, audio and optical flow to generalize to novel classes. We design a cross-modal fusion mechanism to aggregate complimentary multimodal information. Extensive experiments on Kinetic, VGGSound, UCF and HMDB benchmarks demonstrate the effectiveness of our method and the potential of scaling to giant vision and language models.    

\paragraph{Limitations:} We explore three modalities in VGGSound and Kinetics, which does not fully exploit all information available in multimodal videos. In the future, we plan to further improve the model with more modalities like depth and signals from inertial measurement unit (IMU) sensors.

\paragraph{Societal impact:} The proposed method shows better generalization to a wider set of multimodal videos with novel classes, indicating its strong potential for real world applications. We also need to mention that our method is built upon vision and language models pre-trained on large-scale data accumulated automatically from the web with limited manual verification, which may contain biases making them not suitable for some sensitive tasks that or engaging with some social activities.

%%%%%%%%%%%%%%%%%%%%%%%%%%%%%%%%%%%%%%%%%%%%%%%%%%%%%%%%%%%%
\clearpage
\bibliographystyle{plain}
\bibliography{ref}

\begin{thebibliography}{10}

\bibitem{arnab2021vivit}
Anurag Arnab, Mostafa Dehghani, Georg Heigold, Chen Sun, Mario Lu{\v{c}}i{\'c},
  and Cordelia Schmid.
\newblock Vivit: A video vision transformer.
\newblock In {\em ICCV}, 2021.

\bibitem{bishay2019tarn}
Mina Bishay, Georgios Zoumpourlis, and Ioannis Patras.
\newblock Tarn: Temporal attentive relation network for few-shot and zero-shot
  action recognition.
\newblock In {\em BMVC}, 2019.

\bibitem{brattoli2020rethinking}
Biagio Brattoli, Joseph Tighe, Fedor Zhdanov, Pietro Perona, and Krzysztof
  Chalupka.
\newblock Rethinking zero-shot video classification: End-to-end training for
  realistic applications.
\newblock In {\em CVPR}, 2020.

\bibitem{carreira2019short}
Joao Carreira, Eric Noland, Chloe Hillier, and Andrew Zisserman.
\newblock A short note on the kinetics-700 human action dataset.
\newblock {\em arXiv preprint arXiv:1907.06987}, 2019.

\bibitem{carreira2017quo}
Joao Carreira and Andrew Zisserman.
\newblock Quo vadis, action recognition? a new model and the kinetics dataset.
\newblock In {\em CVPR}, 2017.

\bibitem{chen2020vggsound}
Honglie Chen, Weidi Xie, Andrea Vedaldi, and Andrew Zisserman.
\newblock {VGGSound}: A large-scale audio-visual dataset.
\newblock In {\em ICASSP}, 2020.

\bibitem{chen2021elaborative}
Shizhe Chen and Dong Huang.
\newblock Elaborative rehearsal for zero-shot action recognition.
\newblock In {\em ICCV}, 2021.

\bibitem{chen1998audio}
Tsuhan Chen and Ram~R Rao.
\newblock Audio-visual integration in multimodal communication.
\newblock {\em Proceedings of the IEEE}, 1998.

\bibitem{deng2009imagenet}
Jia Deng, Wei Dong, Richard Socher, Li-Jia Li, Kai Li, and Li~Fei-Fei.
\newblock Imagenet: A large-scale hierarchical image database.
\newblock In {\em CVPR}, 2009.

\bibitem{devlin2018bert}
Jacob Devlin, Ming-Wei Chang, Kenton Lee, and Kristina Toutanova.
\newblock Bert: Pre-training of deep bidirectional transformers for language
  understanding.
\newblock In {\em NAACL}, 2019.

\bibitem{dosovitskiy2020image}
Alexey Dosovitskiy, Lucas Beyer, Alexander Kolesnikov, Dirk Weissenborn,
  Xiaohua Zhai, Thomas Unterthiner, Mostafa Dehghani, Matthias Minderer, Georg
  Heigold, Sylvain Gelly, et~al.
\newblock An image is worth 16x16 words: Transformers for image recognition at
  scale.
\newblock In {\em ICLR}, 2021.

\bibitem{elhoseiny2013write}
Mohamed Elhoseiny, Babak Saleh, and Ahmed Elgammal.
\newblock Write a classifier: Zero-shot learning using purely textual
  descriptions.
\newblock In {\em ICCV}, 2013.

\bibitem{fayek2020large}
Haytham~M Fayek and Anurag Kumar.
\newblock Large scale audiovisual learning of sounds with weakly labeled data.
\newblock {\em arXiv preprint arXiv:2006.01595}, 2020.

\bibitem{feichtenhofer2017spatiotemporal}
Christoph Feichtenhofer, Axel Pinz, and Richard~P Wildes.
\newblock Spatiotemporal multiplier networks for video action recognition.
\newblock In {\em CVPR}, 2017.

\bibitem{feichtenhofer2016convolutional}
Christoph Feichtenhofer, Axel Pinz, and Andrew Zisserman.
\newblock Convolutional two-stream network fusion for video action recognition.
\newblock In {\em CVPR}, 2016.

\bibitem{frome2013devise}
Andrea Frome, Greg~S Corrado, Jon Shlens, Samy Bengio, Jeff Dean, Marc'Aurelio
  Ranzato, and Tomas Mikolov.
\newblock Devise: A deep visual-semantic embedding model.
\newblock In {\em NeurIPS}, 2013.

\bibitem{gao2019know}
Junyu Gao, Tianzhu Zhang, and Changsheng Xu.
\newblock I know the relationships: Zero-shot action recognition via two-stream
  graph convolutional networks and knowledge graphs.
\newblock In {\em AAAI}, 2019.

\bibitem{gao2020learning}
Junyu Gao, Tianzhu Zhang, and Changsheng Xu.
\newblock Learning to model relationships for zero-shot video classification.
\newblock In {\em TPAMI}, 2020.

\bibitem{gao2021clip}
Peng Gao, Shijie Geng, Renrui Zhang, Teli Ma, Rongyao Fang, Yongfeng Zhang,
  Hongsheng Li, and Yu~Qiao.
\newblock Clip-adapter: Better vision-language models with feature adapters.
\newblock {\em arXiv preprint arXiv:2110.04544}, 2021.

\bibitem{gemmeke2017audio}
Jort~F Gemmeke, Daniel~PW Ellis, Dylan Freedman, Aren Jansen, Wade Lawrence,
  R~Channing Moore, Manoj Plakal, and Marvin Ritter.
\newblock Audio set: An ontology and human-labeled dataset for audio events.
\newblock In {\em ICASSP}, 2017.

\bibitem{ghiasi2021open}
Golnaz Ghiasi, Xiuye Gu, Yin Cui, and Tsung-Yi Lin.
\newblock Open-vocabulary image segmentation.
\newblock {\em arXiv preprint arXiv:2112.12143}, 2021.

\bibitem{gong21ast}
Yuan Gong, Yu-An Chung, and James Glass.
\newblock Ast: Audio spectrogram transformer.
\newblock In {\em Interspeech}, 2021.

\bibitem{gu2021open}
Xiuye Gu, Tsung-Yi Lin, Weicheng Kuo, and Yin Cui.
\newblock Open-vocabulary object detection via vision and language knowledge
  distillation.
\newblock In {\em ICLR}, 2022.

\bibitem{hadsell2006dimensionality}
Raia Hadsell, Sumit Chopra, and Yann LeCun.
\newblock Dimensionality reduction by learning an invariant mapping.
\newblock In {\em CVPR}, 2006.

\bibitem{hahn2019action2vec}
Meera Hahn, Andrew Silva, and James~M Rehg.
\newblock Action2vec: A crossmodal embedding approach to action learning.
\newblock {\em arXiv preprint arXiv:1901.00484}, 2019.

\bibitem{han2020memory}
Tengda Han, Weidi Xie, and Andrew Zisserman.
\newblock Memory-augmented dense predictive coding for video representation
  learning.
\newblock In {\em ECCV}, 2020.

\bibitem{han2020self}
Tengda Han, Weidi Xie, and Andrew Zisserman.
\newblock Self-supervised co-training for video representation learning.
\newblock In {\em NeurIPS}, 2020.

\bibitem{he2016deep}
Kaiming He, Xiangyu Zhang, Shaoqing Ren, and Jian Sun.
\newblock Deep residual learning for image recognition.
\newblock In {\em CVPR}, 2016.

\bibitem{he2019bag}
Tong He, Zhi Zhang, Hang Zhang, Zhongyue Zhang, Junyuan Xie, and Mu~Li.
\newblock Bag of tricks for image classification with convolutional neural
  networks.
\newblock In {\em CVPR}, 2019.

\bibitem{hershey2017cnn}
Shawn Hershey, Sourish Chaudhuri, Daniel~PW Ellis, Jort~F Gemmeke, Aren Jansen,
  R~Channing Moore, Manoj Plakal, Devin Platt, Rif~A Saurous, Bryan Seybold,
  et~al.
\newblock {CNN} architectures for large-scale audio classification.
\newblock In {\em ICASSP}, 2017.

\bibitem{hinton2015distilling}
Geoffrey Hinton, Oriol Vinyals, Jeff Dean, et~al.
\newblock Distilling the knowledge in a neural network.
\newblock {\em arXiv preprint arXiv:1503.02531}, 2015.

\bibitem{jia2021scaling}
Chao Jia, Yinfei Yang, Ye~Xia, Yi-Ting Chen, Zarana Parekh, Hieu Pham, Quoc Le,
  Yun-Hsuan Sung, Zhen Li, and Tom Duerig.
\newblock Scaling up visual and vision-language representation learning with
  noisy text supervision.
\newblock In {\em ICML}, 2021.

\bibitem{ju2021prompting}
Chen Ju, Tengda Han, Kunhao Zheng, Ya~Zhang, and Weidi Xie.
\newblock Prompting visual-language models for efficient video understanding.
\newblock {\em arXiv preprint arXiv:2112.04478}, 2021.

\bibitem{karpathy2014large}
Andrej Karpathy, George Toderici, Sanketh Shetty, Thomas Leung, Rahul
  Sukthankar, and Li~Fei-Fei.
\newblock Large-scale video classification with convolutional neural networks.
\newblock In {\em CVPR}, 2014.

\bibitem{kay2017kinetics}
Will Kay, Joao Carreira, Karen Simonyan, Brian Zhang, Chloe Hillier, Sudheendra
  Vijayanarasimhan, Fabio Viola, Tim Green, Trevor Back, Paul Natsev, et~al.
\newblock The kinetics human action video dataset.
\newblock {\em arXiv preprint arXiv:1705.06950}, 2017.

\bibitem{kazakos2019epic}
Evangelos Kazakos, Arsha Nagrani, Andrew Zisserman, and Dima Damen.
\newblock Epic-fusion: Audio-visual temporal binding for egocentric action
  recognition.
\newblock In {\em ICCV}, 2019.

\bibitem{kim2021daszl}
Tae~Soo Kim, Jonathan~D Jones, Michael Peven, Zihao Xiao, Jin Bai, Yi~Zhang,
  Weichao Qiu, Alan Yuille, and Gregory~D Hager.
\newblock Daszl: Dynamic action signatures for zero-shot learning.
\newblock In {\em AAAI}, 2021.

\bibitem{kuehne2011hmdb}
Hildegard Kuehne, Hueihan Jhuang, Est{\'\i}baliz Garrote, Tomaso Poggio, and
  Thomas Serre.
\newblock Hmdb: a large video database for human motion recognition.
\newblock In {\em ICCV}, 2011.

\bibitem{lei2015predicting}
Jimmy Lei~Ba, Kevin Swersky, Sanja Fidler, et~al.
\newblock Predicting deep zero-shot convolutional neural networks using textual
  descriptions.
\newblock In {\em ICCV}, 2015.

\bibitem{li2017learning}
Ang Li, Allan Jabri, Armand Joulin, and Laurens Van Der~Maaten.
\newblock Learning visual n-grams from web data.
\newblock In {\em ICCV}, 2017.

\bibitem{li2022language}
Boyi Li, Kilian~Q Weinberger, Serge Belongie, Vladlen Koltun, and Ren{\'e}
  Ranftl.
\newblock Language-driven semantic segmentation.
\newblock In {\em ICLR}, 2022.

\bibitem{lin2022cross}
Chung-Ching Lin, Kevin Lin, Linjie Li, Lijuan Wang, and Zicheng Liu.
\newblock Cross-modal representation learning for zero-shot action recognition.
\newblock In {\em CVPR}, 2022.

\bibitem{lin2019temporal}
Ji~Lin, Chuang Gan, and Song Han.
\newblock Temporal shift module for efficient video understanding.
\newblock In {\em ICCV}, 2019.

\bibitem{liu2021video}
Ze~Liu, Jia Ning, Yue Cao, Yixuan Wei, Zheng Zhang, Stephen Lin, and Han Hu.
\newblock Video swin transformer.
\newblock In {\em CVPR}, 2022.

\bibitem{loshchilov2017decoupled}
Ilya Loshchilov and Frank Hutter.
\newblock Decoupled weight decay regularization.
\newblock {\em arXiv preprint arXiv:1711.05101}, 2017.

\bibitem{mandal2019out}
Devraj Mandal, Sanath Narayan, Sai~Kumar Dwivedi, Vikram Gupta, Shuaib Ahmed,
  Fahad~Shahbaz Khan, and Ling Shao.
\newblock Out-of-distribution detection for generalized zero-shot action
  recognition.
\newblock In {\em CVPR}, 2019.

\bibitem{mazumder2021avgzslnet}
Pratik Mazumder, Pravendra Singh, Kranti~Kumar Parida, and Vinay~P Namboodiri.
\newblock Avgzslnet: Audio-visual generalized zero-shot learning by
  reconstructing label features from multi-modal embeddings.
\newblock In {\em WACV}, 2021.

\bibitem{mercea2022audio}
Otniel-Bogdan Mercea, Lukas Riesch, A~Koepke, and Zeynep Akata.
\newblock Audio-visual generalised zero-shot learning with cross-modal
  attention and language.
\newblock In {\em CVPR}, 2022.

\bibitem{mettes2016imagenet}
Pascal Mettes, Dennis~C Koelma, and Cees~GM Snoek.
\newblock The imagenet shuffle: Reorganized pre-training for video event
  detection.
\newblock In {\em ICMR}, 2016.

\bibitem{mikolov2013efficient}
Tomas Mikolov, Kai Chen, Greg Corrado, and Jeffrey Dean.
\newblock Efficient estimation of word representations in vector space.
\newblock {\em arXiv preprint arXiv:1301.3781}, 2013.

\bibitem{mishra2018generative}
Ashish Mishra, Vinay~Kumar Verma, M~Shiva~Krishna Reddy, S~Arulkumar, Piyush
  Rai, and Anurag Mittal.
\newblock A generative approach to zero-shot and few-shot action recognition.
\newblock In {\em WACV}, 2018.

\bibitem{nagrani2021attention}
Arsha Nagrani, Shan Yang, Anurag Arnab, Aren Jansen, Cordelia Schmid, and Chen
  Sun.
\newblock Attention bottlenecks for multimodal fusion.
\newblock In {\em NeurIPS}, 2021.

\bibitem{oord2018representation}
Aaron van~den Oord, Yazhe Li, and Oriol Vinyals.
\newblock Representation learning with contrastive predictive coding.
\newblock {\em arXiv preprint arXiv:1807.03748}, 2018.

\bibitem{parida2020coordinated}
Kranti Parida, Neeraj Matiyali, Tanaya Guha, and Gaurav Sharma.
\newblock Coordinated joint multimodal embeddings for generalized audio-visual
  zero-shot classification and retrieval of videos.
\newblock In {\em WACV}, 2020.

\bibitem{park2019specaugment}
Daniel~S Park, William Chan, Yu~Zhang, Chung-Cheng Chiu, Barret Zoph, Ekin~D
  Cubuk, and Quoc~V Le.
\newblock Specaugment: A simple data augmentation method for automatic speech
  recognition.
\newblock {\em arXiv preprint arXiv:1904.08779}, 2019.

\bibitem{qian2021spatiotemporal}
Rui Qian, Tianjian Meng, Boqing Gong, Ming-Hsuan Yang, Huisheng Wang, Serge
  Belongie, and Yin Cui.
\newblock Spatiotemporal contrastive video representation learning.
\newblock In {\em CVPR}, 2021.

\bibitem{radford2021learning}
Alec Radford, Jong~Wook Kim, Chris Hallacy, Aditya Ramesh, Gabriel Goh,
  Sandhini Agarwal, Girish Sastry, Amanda Askell, Pamela Mishkin, Jack Clark,
  et~al.
\newblock Learning transferable visual models from natural language
  supervision.
\newblock In {\em ICML}, 2021.

\bibitem{simonyan2014two}
Karen Simonyan and Andrew Zisserman.
\newblock Two-stream convolutional networks for action recognition in videos.
\newblock In {\em NeurIPS}, 2014.

\bibitem{smith2005development}
Linda Smith and Michael Gasser.
\newblock The development of embodied cognition: Six lessons from babies.
\newblock {\em Artificial life}, 2005.

\bibitem{soomro2012ucf101}
Khurram Soomro, Amir~Roshan Zamir, and Mubarak Shah.
\newblock Ucf101: A dataset of 101 human actions classes from videos in the
  wild.
\newblock {\em arXiv preprint arXiv:1212.0402}, 2012.

\bibitem{szegedy2015going}
Christian Szegedy, Wei Liu, Yangqing Jia, Pierre Sermanet, Scott Reed, Dragomir
  Anguelov, Dumitru Erhan, Vincent Vanhoucke, and Andrew Rabinovich.
\newblock Going deeper with convolutions.
\newblock In {\em CVPR}, 2015.

\bibitem{tran2015learning}
Du~Tran, Lubomir Bourdev, Rob Fergus, Lorenzo Torresani, and Manohar Paluri.
\newblock Learning spatiotemporal features with 3d convolutional networks.
\newblock In {\em ICCV}, 2015.

\bibitem{tran2018closer}
Du~Tran, Heng Wang, Lorenzo Torresani, Jamie Ray, Yann LeCun, and Manohar
  Paluri.
\newblock A closer look at spatiotemporal convolutions for action recognition.
\newblock In {\em CVPR}, 2018.

\bibitem{vaswani2017attention}
Ashish Vaswani, Noam Shazeer, Niki Parmar, Jakob Uszkoreit, Llion Jones,
  Aidan~N Gomez, {\L}ukasz Kaiser, and Illia Polosukhin.
\newblock Attention is all you need.
\newblock In {\em NeurIPS}, 2017.

\bibitem{wang2016temporal}
Limin Wang, Yuanjun Xiong, Zhe Wang, Yu~Qiao, Dahua Lin, Xiaoou Tang, and
  Luc~Van Gool.
\newblock Temporal segment networks: Towards good practices for deep action
  recognition.
\newblock In {\em ECCV}, 2016.

\bibitem{wang2021actionclip}
Mengmeng Wang, Jiazheng Xing, and Yong Liu.
\newblock Actionclip: A new paradigm for video action recognition.
\newblock {\em arXiv preprint arXiv:2109.08472}, 2021.

\bibitem{wang2017zero}
Qian Wang and Ke~Chen.
\newblock Zero-shot visual recognition via bidirectional latent embedding.
\newblock {\em IJCV}, 2017.

\bibitem{wu2016harnessing}
Zuxuan Wu, Yanwei Fu, Yu-Gang Jiang, and Leonid Sigal.
\newblock Harnessing object and scene semantics for large-scale video
  understanding.
\newblock In {\em CVPR}, 2016.

\bibitem{xiao2020audiovisual}
Fanyi Xiao, Yong~Jae Lee, Kristen Grauman, Jitendra Malik, and Christoph
  Feichtenhofer.
\newblock Audiovisual slowfast networks for video recognition.
\newblock {\em arXiv preprint arXiv:2001.08740}, 2020.

\bibitem{xie2018rethinking}
Saining Xie, Chen Sun, Jonathan Huang, Zhuowen Tu, and Kevin Murphy.
\newblock Rethinking spatiotemporal feature learning: Speed-accuracy trade-offs
  in video classification.
\newblock In {\em ECCV}, 2018.

\bibitem{xu2017transductive}
Xun Xu, Timothy Hospedales, and Shaogang Gong.
\newblock Transductive zero-shot action recognition by word-vector embedding.
\newblock {\em IJCV}, 2017.

\bibitem{xu2016multi}
Xun Xu, Timothy~M Hospedales, and Shaogang Gong.
\newblock Multi-task zero-shot action recognition with prioritised data
  augmentation.
\newblock In {\em ECCV}, 2016.

\bibitem{yu2022coca}
Jiahui Yu, Zirui Wang, Vijay Vasudevan, Legg Yeung, Mojtaba Seyedhosseini, and
  Yonghui Wu.
\newblock Coca: Contrastive captioners are image-text foundation models.
\newblock {\em arXiv preprint arXiv:2205.01917}, 2022.

\bibitem{yuan2021florence}
Lu~Yuan, Dongdong Chen, Yi-Ling Chen, Noel Codella, Xiyang Dai, Jianfeng Gao,
  Houdong Hu, Xuedong Huang, Boxin Li, Chunyuan Li, et~al.
\newblock Florence: A new foundation model for computer vision.
\newblock {\em arXiv preprint arXiv:2111.11432}, 2021.

\bibitem{zach2007duality}
Christopher Zach, Thomas Pock, and Horst Bischof.
\newblock A duality based approach for realtime tv-l 1 optical flow.
\newblock In {\em PR}, 2007.

\bibitem{zhang2018visual}
Chenrui Zhang and Yuxin Peng.
\newblock Visual data synthesis via {GAN} for zero-shot video classification.
\newblock In {\em IJCAI}, 2018.

\bibitem{zhou2021coop}
Kaiyang Zhou, Jingkang Yang, Chen~Change Loy, and Ziwei Liu.
\newblock Learning to prompt for vision-language models.
\newblock {\em arXiv preprint arXiv:2109.01134}, 2021.

\bibitem{zhou2022cocoop}
Kaiyang Zhou, Jingkang Yang, Chen~Change Loy, and Ziwei Liu.
\newblock Conditional prompt learning for vision-language models.
\newblock In {\em CVPR}, 2022.

\bibitem{zhu2018towards}
Yi~Zhu, Yang Long, Yu~Guan, Shawn Newsam, and Ling Shao.
\newblock Towards universal representation for unseen action recognition.
\newblock In {\em CVPR}, 2018.

\end{thebibliography}

%%%%%%%%%%%%%%%%%%%%%%%%%%%%%%%%%%%%%%%%%%%%%%%%%%%%%%%%%%%%
\clearpage
\appendix
\section{Appendix}
\label{appx:temperature_tuning}

\paragraph{Temperature tuning.} As described in Sec.~\ref{subsec:openset}, in addition to fused flow and audio features of $\{\mathbf{f_m}, \mathbf{a_m}\}$, we also incorporate the video feature
$\mathbf{v}$ extracted from the frozen video backbone to enhance the generalization to novel classes. We denote the probability distribution followed by $\{p_{f}(j) |_{j=1}^q\}$, $\{p_{a}(j) |_{j=1}^q\}$ and $\{p_{v}(j) |_{j=1}^q\}$ as $D_f$, $D_a$ and $D_v$. 
In our experiments we find the curve of $D_v$ tends to be much flatter (or have higher information entropy) than $D_f$ and $D_a$ when the temperatures $\tau_v$, $\tau_f$ and $\tau_a$ are all set to the CLIP's default value of 0.01. Neglecting this difference and directly combining the scores as in Eq.~\ref{eq:oset_fusion} would lead to poor performance.
We address this problem by lowering $\tau_v$ so that the distribution of $D_v$ would be more similar to $D_f$ and $D_a$. As shown in Tab.~\ref{tab:ablation_calibration}, adjusting $\tau_v$ to 0.003 while keeping $\tau_f$ and $\tau_a$ as 0.01 greatly improves the performance by 20\% on Kinetics-700 and 16\% on VGGSound.

\begin{table}[hbtp]
\begin{minipage}[t]{0.45\textwidth}
\centering
\begin{tabular}{cccc}
    \toprule 
    $\mathbf{v}$ acc. & $\mathbf{f_m}$ acc. & $\tau_v$ & Final acc. \\
    \midrule 
    56.7 & 30.4 & 0.01 & 38.0 \\
    \hdashline
    56.7 & 30.4 & 0.003 & \textbf{58.1} \\
    56.7 & 30.4 & 0.001 & 57.1 \\
    56.7 & 30.4 & 0.0003 & 56.0 \\
    56.7 & 30.4 & 0.0001 & 56.4 \\
    \bottomrule 
\end{tabular}
\end{minipage}
\hfill
\begin{minipage}[t]{0.45\textwidth}
\centering
\begin{tabular}{cccc}
    \toprule 
    $\mathbf{v}$ acc. & $\mathbf{a_m}$ acc. & $\tau_v$ & Final acc. \\
    \midrule 
    48.8 & 24.8 & 0.01 & 35.7\\
    \hdashline
    48.8 & 24.8 & 0.003 & \textbf{51.5} \\
    48.8 & 24.8 & 0.001 & 49.5 \\
    48.8 & 24.8 & 0.0003 & 49.1 \\
    48.8 & 24.8 & 0.0001 & 49.0 \\
    \bottomrule 
\end{tabular}
\end{minipage}

\begin{minipage}[b]{0.45\textwidth}
\vspace{3pt}
\hspace{-35pt}
\centering
(a) \textbf{Tuning $\tau_v$ on Kinectics-700.}
\end{minipage}
\begin{minipage}[b]{0.45\textwidth}
\vspace{3pt}
\hspace{20pt}
\centering
(b) \textbf{Tuning $\tau_v$ on VGGSound.}
\end{minipage}

\caption{\textbf{Ablation on temperature tuning}. Compared with using CLIP's default temperature of 0.01 (the first row), using a smaller temperature of 0.003 could greatly improve the performance by 20\% on Kinetics-700 and 16\% on VGGSound.}
\label{tab:ablation_calibration}
\end{table}

\end{document}